\documentclass[11pt]{article}
\usepackage[utf8]{inputenc}
\usepackage[T1]{fontenc}
\usepackage[margin=1in]{geometry}
\usepackage{amsmath}
\usepackage{booktabs}
\usepackage{graphicx}
\usepackage{hyperref}
\usepackage{natbib}
\usepackage{xcolor}
\usepackage{listings}

\title{evalci: A Python Library for Statistically Rigorous\\ Comparison of Language Model Evaluations}
\author{Shreyas K Chandrahas}
\date{}

\begin{document}
\maketitle

\begin{abstract}
The dominant practice in language model evaluation is to report a single accuracy
number per model and declare the higher one better, without testing whether the
gap could plausibly be sampling noise. On benchmarks of a few thousand items, and
under temperature sampling where a model can differ from itself run to run by more
than the reported gap between models, this practice routinely overstates
confidence in headline claims. The statistical machinery to fix this --
confidence intervals, paired significance tests, power analysis, clustered
standard errors, multiple-comparison correction -- is well established, but no
standard, pip-installable tool packages it in the shape an evaluation actually
takes: a per-item results table. We present \texttt{evalci}, a pure-Python
library (numpy/scipy/pandas only) that turns a per-item results table into a
publication-ready claim -- e.g., ``Model A beats Model B, $\Delta=3.1$ pts, 95\%
CI [1.2, 5.0], paired permutation $p=0.002$, $n=1{,}319$'' -- in one function
call, with adapters for \texttt{lm-evaluation-harness} and HELM output. Every
routine is validated against an independent reference (\texttt{statsmodels},
or brute-force exact enumeration) rather than only against itself. As a case
study, we re-analyze a public comparison of nine language models' MMLU accuracy
and find that 3 of the 8 adjacent leaderboard-rank gaps are not statistically
significant after correcting for the 36 pairwise comparisons the ranking
implies. \texttt{evalci} is available at
\url{https://pypi.org/project/evalci/} (source:
\url{https://github.com/Shreyaskc/evalci}, DOI:
\url{https://doi.org/10.5281/zenodo.21201815}).
\end{abstract}

\section{Introduction}

A language model evaluation is an experiment: a fixed set of items is scored
against a metric, and the result is a noisy estimate of an underlying
population quantity, not the quantity itself. The evaluation literature has
largely treated it as something else -- a deterministic measurement -- and
reported the resulting number without an error bar. This is not a cosmetic
omission. On a benchmark of a few thousand items, a 2-point gap between two
models' accuracy is frequently within the noise implied by the sample size
alone. Under temperature sampling, the same model evaluated twice can differ
from itself by more than the reported gap between two different models; recent
work quantifying this shows that properly clustered standard errors -- which
account for repeated sampling of the same item -- can be more than 3x larger
than the naive standard errors typically reported \citep{miller2024}.

This is not a new observation. \citet{dror2018} surveyed significance testing
practice in ACL and TACL papers and found it routinely ignored or misapplied.
\citet{card2020} showed that many standard NLP test sets are underpowered for
the effect sizes papers actually claim. \citet{miller2024} derived the
statistical machinery needed to analyze evaluations properly -- clustered
standard errors, power calculations -- as a methods note. What has been
missing is not the statistics but a standard, tested, pip-installable
implementation shaped like an evaluation: a table of \emph{items}, not a pair
of scalars. \texttt{scipy} and \texttt{statsmodels} provide the general-purpose
primitives (a proportion confidence interval, a bootstrap, a multiple-testing
correction), but none of them speak the vocabulary an evaluation actually
produces -- per-item, paired-by-question, clustered-by-repeated-decode,
many-models-by-many-benchmarks -- and none of them read
\texttt{lm-evaluation-harness} or HELM output directly.

We present \texttt{evalci}, which closes that gap. Its contribution is not a
new statistical method; every routine it implements is decades old. Its
contribution is packaging: a small, tested, pure-Python library, installable
with \texttt{pip} in under five minutes with no GPU, that takes the exact
per-item table an evaluation harness already produces and returns a
publication-ready confidence interval, significance test, or corrected
many-model comparison table. Section~\ref{sec:design} describes the library.
Section~\ref{sec:methods} describes the statistical methods it implements.
Section~\ref{sec:validation} describes how each was validated against an
independent reference rather than only against itself. Section~\ref{sec:case-study}
uses \texttt{evalci} to re-analyze a public multi-model comparison and shows
that a third of the adjacent leaderboard-rank gaps in it are not statistically
significant once the comparisons actually being made are corrected for.

\section{Library Design}
\label{sec:design}

\texttt{evalci} depends only on \texttt{numpy}, \texttt{scipy}, and
\texttt{pandas} \citep{harris2020numpy,virtanen2020scipy,mckinney2010pandas},
so \texttt{pip install evalci} works on a laptop with no GPU. The library is
built around a single shared contract: a per-item results table with columns
\texttt{item\_id}, \texttt{model}, \texttt{score}, and optionally
\texttt{subset} (for stratifying by benchmark or category) and
\texttt{sample\_idx} (for repeated decodes of the same item). Every function,
and every adapter, produces or consumes this schema, so results loaded from an
\texttt{lm-evaluation-harness} run, a HELM run, or a plain CSV compose freely.

The public API is six functions:

\begin{lstlisting}
evalci.ci(scores, method="wilson"|"clopper-pearson"|"bootstrap")
evalci.compare(a, b, paired=True,
               method="permutation"|"bootstrap"|"mcnemar")
evalci.power(delta, n=None, power=0.8,
             method="analytic"|"simulation")
evalci.multi_compare(df, correction="holm"|"bh")
evalci.cluster_ci(scores, clusters)
evalci.report(result, format="markdown"|"latex")
\end{lstlisting}

\noindent A typical call looks like:

\begin{lstlisting}
>>> result = evalci.compare(model_a_scores, model_b_scores,
...                          method="permutation")
>>> evalci.report(result)
'Δ=0.034, 95% CI [0.005, 0.063], paired permutation p=0.025*, n=1319'
\end{lstlisting}

Adapters (\texttt{evalci.adapters.load\_lm\_eval\_harness},
\texttt{load\_helm}, \texttt{load\_csv}) parse per-item output from
\texttt{lm-evaluation-harness} and HELM directly into the shared schema, and a
CLI (\texttt{evalci compare results\_a.json results\_b.json}) wraps the common
case of comparing two result files without writing any Python.

\section{Statistical Methods}
\label{sec:methods}

\subsection{Confidence intervals}
For a single model's binary (correct/incorrect) per-item scores,
\texttt{evalci.ci} implements the Wilson score interval
\citep{wilson1927} and the exact Clopper-Pearson interval
\citep{clopperpearson1934}. For continuous scores, or when the binomial
assumption is not appropriate, it wraps \texttt{scipy}'s bootstrap
implementation \citep{virtanen2020scipy} to provide percentile and
bias-corrected-and-accelerated (BCa) intervals \citep{efron1979,efrontibshirani1993}.

\subsection{Paired and unpaired comparison}
\texttt{evalci.compare} implements three significance tests. The paired
\emph{sign-flip permutation test} randomly flips the sign of each per-item
difference under the null that the two models are exchangeable on that item;
for $n\le12$ items it enumerates all $2^n$ sign patterns exactly rather than
approximating with Monte Carlo. The \emph{null-shifted bootstrap hypothesis
test} centers the observed differences at zero and resamples with
replacement, giving a nonparametric alternative to the permutation test with
the same asymptotic behavior. \emph{McNemar's test} \citep{mcnemar1947},
exact (binomial) below $n=25$ discordant pairs and asymptotic
(chi-squared, with continuity correction) above, tests paired binary outcomes
directly via the discordant-pair counts. All three paired tests are
accompanied by a bootstrap confidence interval on the mean paired difference,
so a single \texttt{compare()} call returns both a significance test and an
interval estimate. Unpaired variants (label-shuffle permutation, an
independent-samples null-shifted bootstrap) are available when items are not
aligned across the two models being compared -- the situation in
Section~\ref{sec:case-study}, where only aggregate accuracy, not per-item
correctness, is public.

\subsection{Multiple-comparison correction}
\texttt{evalci.multi\_compare} runs every pairwise model comparison
(optionally stratified by benchmark subset) and applies Holm's step-down
procedure \citep{holm1979} or Benjamini-Hochberg FDR control
\citep{benjaminihochberg1995} across \emph{all} of them -- not just the
subset a user might otherwise be tempted to report -- before returning a
table with adjusted $p$-values and a significance flag.

\subsection{Clustered standard errors}
\texttt{evalci.cluster\_ci} resamples whole clusters, rather than individual
observations, in its bootstrap -- the correct treatment when items are not
independent, as in repeated decodes of the same question under temperature
sampling, or grouped items sharing an underlying passage. This directly
addresses the clustering problem quantified by \citet{miller2024}: treating
repeated samples as independent observations understates the true standard
error.

\subsection{Power analysis}
\texttt{evalci.power} answers ``how many items do I need to detect a
$\delta$-point gap at 80\% power?'' (or, given $n$, what power is achieved) via
a closed-form two-proportion normal approximation, with a Monte Carlo
simulation fallback (\texttt{method="simulation"}) that accepts a correlation
parameter $\rho$ -- the fraction of items whose correctness is linked across
the two models by shared difficulty rather than drawn independently -- for
when items are not independent across models. This is motivated directly by
\citet{card2020}'s finding that standard-sized NLP test sets are underpowered
for many effect sizes papers report as significant.

\section{Validation}
\label{sec:validation}

Statistical correctness is the point of this library, so every routine is
checked against an independent reference rather than only re-tested against
its own implementation:

\begin{itemize}
  \item Wilson and Clopper-Pearson intervals are checked against
  \texttt{statsmodels}'s \texttt{proportion\_confint}
  \citep{seabold2010statsmodels} across a range of $(k, n)$.
  \item McNemar's test, exact and asymptotic, is checked against
  \texttt{statsmodels}'s \texttt{mcnemar}.
  \item Holm and Benjamini-Hochberg correction are checked against
  \texttt{statsmodels}'s \texttt{multipletests}.
  \item The paired permutation test is checked against brute-force exact
  enumeration of all $2^n$ sign-flip patterns for small $n$, and Monte Carlo
  estimates are shown to converge to the exact value as the resample count
  grows.
  \item Bootstrap confidence intervals are checked with a coverage
  simulation: over 200 independent synthetic trials, a nominal 95\% CI
  contains the true parameter in the expected range of trials.
\end{itemize}

\texttt{statsmodels} is a test-only dependency (installed via
\texttt{pip install evalci[test]}), not a runtime dependency of the installed
package -- it is used only to validate \texttt{evalci}'s own,
dependency-free implementations. At the time of writing, the test suite
comprises 100 tests and covers 92--99\% of the statistical core
(\texttt{\_intervals.py}, \texttt{\_significance.py}, \texttt{\_correction.py},
\texttt{\_power.py}, \texttt{stats.py}); continuous integration runs the full
suite on every push across Python 3.9--3.12.

\section{Case Study: How Many Adjacent Leaderboard Ranks Are Actually Different?}
\label{sec:case-study}

As a concrete demonstration -- and the kind of re-analysis \texttt{evalci} is
meant to make routine -- we ask how many of the rank differences implied by a
public multi-model comparison table survive a significance test.

\subsection{Data}
We use Table 2 of \citet{grattafiori2024}, ``Performance of finetuned Llama 3
models on key benchmark evaluations,'' which reports 5-shot MMLU accuracy for
nine instruction-tuned models evaluated in a single table: Llama 3 8B, GPT-3.5
Turbo, Mixtral 8x22B, Nemotron-4 340B, Llama 3 70B, GPT-4 (0125), Llama 3 405B,
GPT-4o, and Claude 3.5 Sonnet, with accuracies ranging from 69.4\% to 89.9\%.
The MMLU test set contains $n=14{,}042$ questions \citep{hendrycks2021}.

\subsection{Method}
Per-item correctness for these models is not public -- only the aggregate
accuracy is. We reconstruct, for each model, an i.i.d.\ binary vector of
length $n$ consistent with its reported accuracy ($k=\mathrm{round}(\text{accuracy}\times n)$
ones, the remainder zeros) and compare every pair with \texttt{multi\_compare}
using \texttt{correction="holm"}, \texttt{method="permutation"}, and
\texttt{paired=False}. This is deliberately conservative: true per-item alignment,
were it public, would let us run a \emph{paired} test, and under the
realistic assumption that item difficulty is positively correlated across
models (some MMLU questions are simply harder for every model), a paired test
is more powerful than the unpaired test used here. The count of
non-significant gaps we report below is therefore an upper bound -- a true
paired re-analysis could only find \emph{fewer} non-significant adjacent
gaps, not more.

\subsection{Result}
\begin{figure}[htbp]
  \centering
  \includegraphics[width=0.85\textwidth]{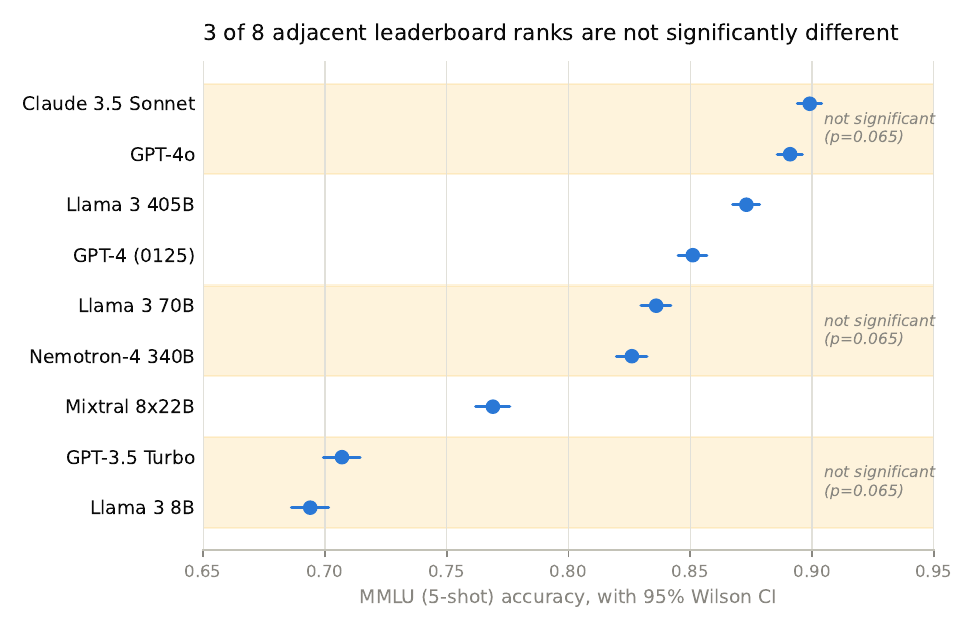}
  \caption{MMLU (5-shot) accuracy for nine publicly reported models, with 95\%
  Wilson confidence intervals reconstructed from aggregate accuracy and
  $n=14{,}042$. Shaded bands mark the three adjacent-rank pairs whose gap is
  not statistically significant ($p_{\mathrm{adj}}>0.05$) after Holm
  correction across all 36 pairwise comparisons.}
  \label{fig:leaderboard}
\end{figure}

\texttt{multi\_compare} evaluates all $\binom{9}{2}=36$ pairwise comparisons
and Holm-corrects across all of them, not just the eight adjacent-rank pairs
we highlight -- the correct scope for the correction, since failing to account
for comparisons actually performed (even ones not reported) is itself a
common significance-testing error \citep{dror2018}. Figure~\ref{fig:leaderboard}
shows the resulting 95\% Wilson confidence intervals for each model, with the
three adjacent-rank pairs that are \emph{not} significant at $\alpha=0.05$
after correction shaded.

\begin{table}[htbp]
\centering
\small
\begin{tabular}{llrrl}
\toprule
Model A & Model B & $\Delta$ (pts) & $p_{\mathrm{adj}}$ & Significant? \\
\midrule
GPT-3.5 Turbo & Llama 3 8B & 1.3 & 0.065 & No \\
Mixtral 8x22B & GPT-3.5 Turbo & 6.2 & 0.0036 & Yes \\
Nemotron-4 340B & Mixtral 8x22B & 5.7 & 0.0036 & Yes \\
Llama 3 70B & Nemotron-4 340B & 1.0 & 0.065 & No \\
GPT-4 (0125) & Llama 3 70B & 1.5 & 0.0036 & Yes \\
Llama 3 405B & GPT-4 (0125) & 2.2 & 0.0036 & Yes \\
GPT-4o & Llama 3 405B & 1.8 & 0.0036 & Yes \\
Claude 3.5 Sonnet & GPT-4o & 0.8 & 0.065 & No \\
\bottomrule
\end{tabular}
\caption{All 8 adjacent-rank pairs from the ranked MMLU comparison, Holm-adjusted
$p$-values from the full 36-comparison family. 3 of 8 are not significant.}
\label{tab:adjacent}
\end{table}

Three of the eight adjacent-rank gaps -- GPT-3.5 Turbo vs.\ Llama 3 8B,
Nemotron-4 340B vs.\ Llama 3 70B, and GPT-4o vs.\ Claude 3.5 Sonnet -- are not
statistically distinguishable at $\alpha=0.05$ once the correction accounts
for the full set of comparisons the ranking implies. Notably, all three land
on the identical adjusted $p$-value (0.065): Holm's step-down procedure
enforces monotonicity of adjusted $p$-values in sorted order, so once a larger
adjusted value appears it propagates forward to subsequent, nominally smaller,
raw $p$-values -- a real property of the correction, not an artifact, and a
concrete illustration of why Holm-adjusted $p$-values should be read as a
corrected significance threshold rather than a precise per-comparison
estimate. The full 36-comparison table, and the script that reproduces it
end-to-end from the raw accuracy figures, are included with this paper's
source.

\section{Limitations}

The case study reconstructs i.i.d.\ per-item vectors from aggregate accuracy
because true per-item data for these models is not public; this preserves
the point estimate and lets us bound significance conservatively
(Section~\ref{sec:case-study}), but it cannot recover any real correlation
structure between items, and a genuine per-item re-analysis -- exactly what
\texttt{evalci}'s \texttt{lm-evaluation-harness} and HELM adapters are for --
would be preferable wherever per-item logs are available. Separately, the
source table itself does not disclose whether every competitor number was
reproduced under one evaluation harness or drawn from each vendor's own
report; \texttt{evalci} computes correctly conditional on the numbers given
to it, but cannot correct for evaluation-protocol inconsistency upstream of
those numbers. Monte Carlo permutation and bootstrap $p$-values have finite
resolution ($1/(n_{\mathrm{resamples}}+1)$); very small nominal $p$-values
should use a larger resample count. Wilson and Clopper-Pearson intervals
assume independent binary per-item scores; correlated or continuous scores
should use \texttt{cluster\_ci} or the bootstrap methods instead. The power
simulation's correlation parameter $\rho$ is a simplified shared-difficulty
knob, not a fitted bivariate model, and should be treated as indicative.

\section{Availability}

\texttt{evalci} is released under the MIT license and available via
\texttt{pip install evalci} (\url{https://pypi.org/project/evalci/}). Source,
issue tracker, and the reproducible case-study scripts are at
\url{https://github.com/Shreyaskc/evalci}. Continuous integration runs the
test suite across Python 3.9--3.12 on every push. Tagged releases are archived
on Zenodo (concept DOI \url{https://doi.org/10.5281/zenodo.21201815}).

\section{Conclusion}

Reviewers increasingly ask evaluation papers for error bars; until now,
authors who wanted to answer correctly had to reimplement paired bootstrap
and permutation tests by hand, so most did not. \texttt{evalci} packages
decades-old, individually well-validated statistical methods into the one
shape an evaluation actually takes -- a per-item table -- and validates the
implementation against an independent reference rather than only against
itself. The case study in Section~\ref{sec:case-study} is not a special case:
applied to any public leaderboard, the same one-line \texttt{multi\_compare}
call will show how many of its rank differences are noise. We hope making
that call this cheap raises the floor of the field's empirical rigor the way
general-purpose scientific-computing libraries have for other parts of the
stack.

\bibliographystyle{plainnat}
\bibliography{references}

\end{document}